\documentclass[runningheads]{llncs}
\usepackage[T1]{fontenc}
\usepackage{graphicx}
\usepackage{cite}
\usepackage{amsmath}
\usepackage{array}
\usepackage[dvipsnames, table]{xcolor}
\usepackage{orcidlink}
\usepackage{pdfpages}
\usepackage{bbm}
\usepackage{multirow}
\usepackage{soul} % for \st{}  (strike through)

\DeclareMathOperator*{\argmin}{argmin}

\begin{document}

\title{Integrating Temporal Context into Streaming Data for Human Activity Recognition in Smart Home\thanks{This study is conducted as part of Innovate UK Knowledge Transfer Partnership 13562. The financial support from Legrand Care and Innovate UK is gratefully acknowledged.}}
\titlerunning{Integrating Temporal Context into Streaming Data}
\author{Marina~Vicini\inst{1, 2}\orcidlink{0009-0006-5622-4216} 
\and Martin~Rudorfer\inst{1}\orcidlink{0000-0001-9109-5188}
\and Zhuangzhuang~Dai\inst{1}\orcidlink{0000-0002-6098-115X}
\and Luis~J.~Manso\inst{1}\orcidlink{0000-0003-2616-1120}}
\authorrunning{M. Vicini et al.}
\institute{Aston University, Aston St, Birmingham B4 7ET, United Kingdom \\
\and Legrand Care, Innovation House, Northumberland Business Park, 5 Silverton court, Cramlington NE23 7RY, United Kingdom \\
\email{\{m.vicini, m.rudorfer, z.dai1, l.manso\}@aston.ac.uk}}
\maketitle              % typeset the header of the contribution
\begin{abstract}
% The abstract should briefly summarize the contents of the paper in 150--250 words.
 
With the global population ageing, it is crucial to enable individuals to live independently and safely in their homes. 
Using ubiquitous sensors such as Passive InfraRed sensors (PIR) and door sensors is drawing increasing interest for monitoring daily activities and facilitating preventative healthcare interventions for the elderly.
Human Activity Recognition (HAR) from passive sensors mostly relies on traditional machine learning and includes data segmentation, feature extraction, and classification.
While techniques like Sensor Weighting Mutual Information (SWMI) capture spatial context in a feature vector, effectively leveraging temporal information remains a challenge.
% contribution
We tackle this by clustering activities into morning, afternoon, and night, and encoding them into the feature weighting method calculating distinct mutual information matrices.
We further propose to extend the feature vector by incorporating time of day and day of week as cyclical temporal features, as well as adding a feature to track the user's location.
The experiments show improved accuracy and F1-score over existing state-of-the-art methods in three out of four real-world datasets, with highest gains in a low-data regime.
These results highlight the potential of our approach for developing effective smart home solutions to support ageing in place.

\end{abstract}

\section{Introduction}
The global population is rapidly aging, with an estimated 1.6 billion individuals aged over 65 by 2050 \cite{un-data-2022}. This age group is growing faster than others: the percentage of the population over 65 is expected to rise from 10\% in 2022 to 16\% in 2050. 
As people age, they often experience declines in physical and cognitive abilities, raising the risk of chronic diseases and dependence on others for everyday tasks. These changes frequently necessitate adapting their living environment or relocating~\cite{perry2014relocation}. 
The World Health Organization highlights that retaining the ability to choose is crucial for self-determination, independence, and dignity~\cite{world2015world}.
Allowing individuals to live independently for as long as possible, offers significant benefits for autonomy and reduces healthcare costs~\cite{golant2008commentary, marek2012aging}.

\begin{figure}[!t]
    \centering
    \includegraphics[trim=10 58 45 12,clip, width=10cm]{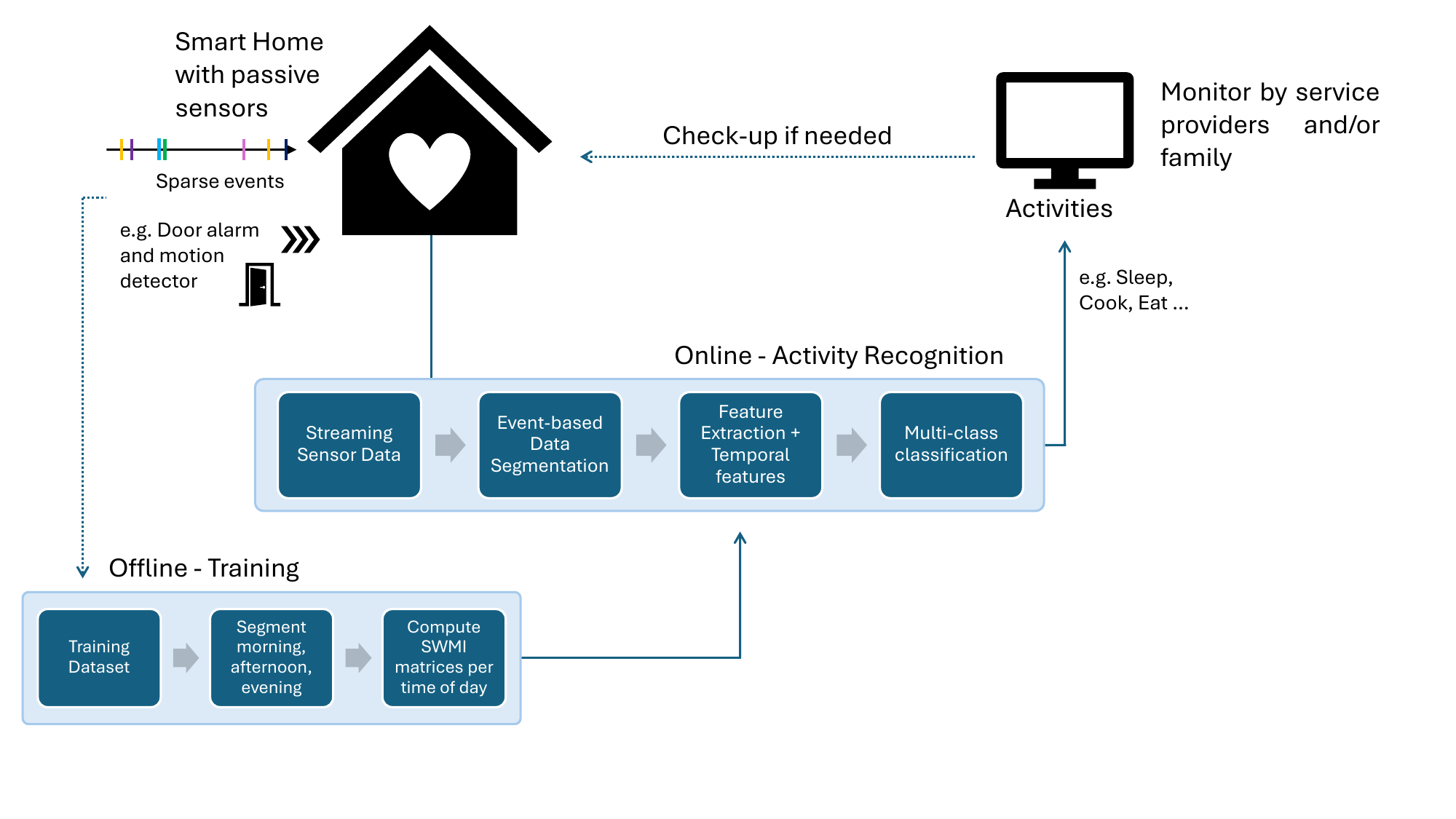}
    \caption{Data is acquired through devices in the users' smart homes. After segmenting the data, features are extracted and weighted with temporal Mutual Information matrices. Lastly, a multi-class classifier is employed. Activities can be monitored by service providers or family.}
    \label{fig:process}
\end{figure}

Facilitating aging in place requires technologies and services to manage individuals' health and daily lives.
Ambience technology can assess a person's ability to perform Activities of Daily Living (ADL), an indicator of their functional status \cite{edemekong2019activities}. 
Human Activity Recognition (HAR), aims to infer activities from sensor data, and can support timely assistance, context-aware prompts~\cite{das2011automated, das2012prompt}, and monitor routine habits to detect changes that may indicate health issues, allowing for proactive intervention~\cite{deep2019survey, wang2023survey}, as illustrated in Figure~\ref{fig:process}.

Vision-based, wearable, contact-based, and Passive InfraRed (PIR) sensors are common sensor types used by HAR systems.
Vision offers rich data but faces privacy concerns~\cite{dang2020sensor, claes2015attitudes}. 
Wearable sensors are popular due to their accuracy and real-time monitoring but are impractical for long-term use due to comfort and maintenance issues~\cite{zhang2022deep, xu2022online}.
On the other hand, PIRs and contact-based sensors are less intrusive, in addition to being cost-effective and low-maintenance.
Notable projects using these sensors for HAR in smart homes include CASAS~\cite{cook2012casas} and PlaceLab~\cite{intille2005living}. 

A remarkable difference when working with these sensors, as opposed to cameras or wearables, is that the resulting stream of data consists of sparse events with irregular timings (\textit{e.g.}, open/close for door sensors).
The events are typically grouped into segments and aggregated into feature vectors which are passed on as input for classic machine learning algorithms.
While some approaches incorporate spatial cues, such as the co-occurrence of sensor events, current feature extraction methods neglect most of the temporal cues.
For example, in most methods, the order of sensor events is not represented in the feature vector.
We directly address these shortcomings in our work.

Our main contributions improve the expressiveness of the features by integrating temporal context in several ways:
\begin{itemize}
    \item we cluster activities into morning, afternoon, and night, and encode this into the feature weighting method based on mutual sensor information by~\cite{krishnan2014activity};
    \item we extend the feature vector by adding cyclical temporal features which represent the time of day and day of the week;
    \item we extend the feature vector by encoding location changes that precede the latest sensor event.
\end{itemize}
The experimental evaluation on four different datasets from the CASAS project \cite{cook2012casas} includes ablation studies to identify the improvement stemming from each contribution, and a comparison against existing state-of-the-art methods.
Our experiments demonstrate the usefulness of incorporating temporal context into the features.
The rest of the paper is divided as follows: Section \ref{sec:background} discusses different techniques to segment streaming data, Section \ref{sec:methodology} introduces our approach, and experimental results are presented in Section \ref{sec:experiments}. Conclusion and future works are found in Section \ref{sec:conclusion}.

\section{Related Works}  \label{sec:background}
\begin{figure}[h]
    \centering
    \includegraphics[trim=50 180 160 50,clip,width=11cm]
    {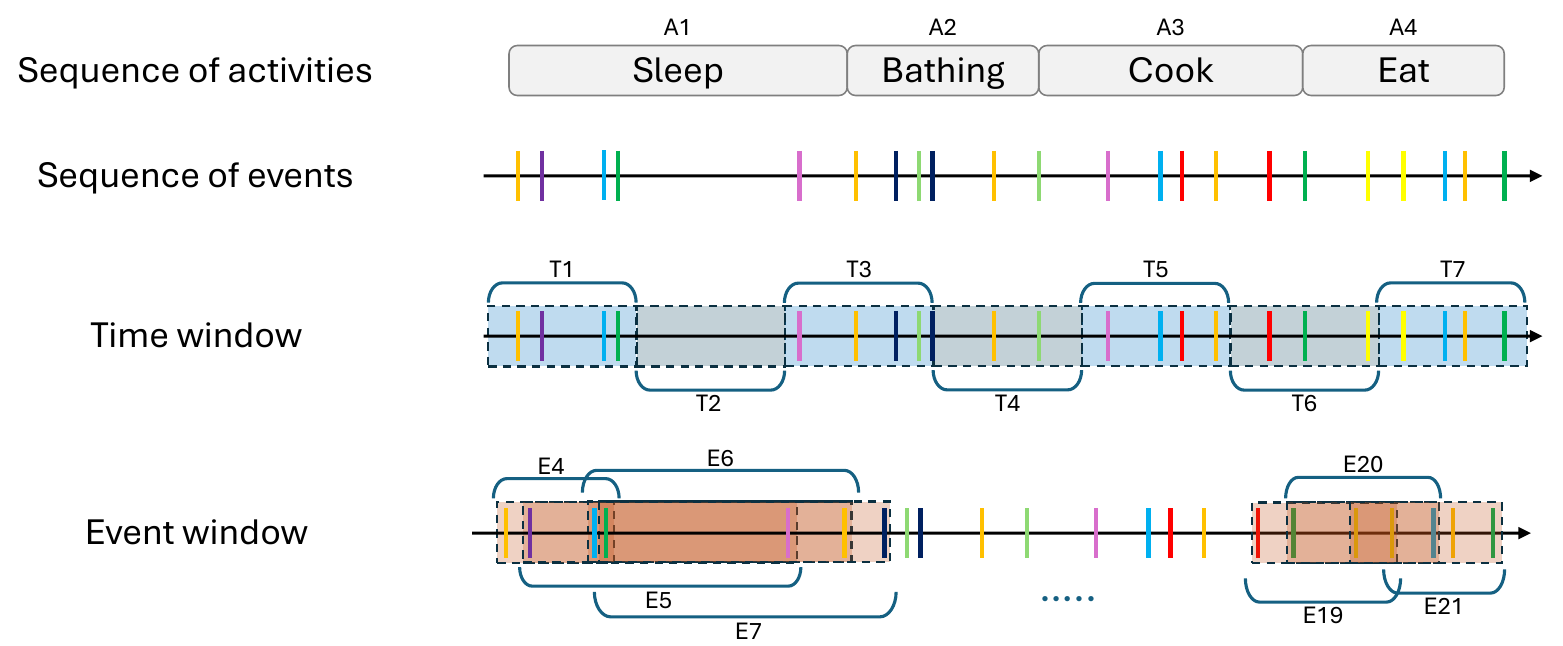}
    \caption{Different data segmentation techniques. Colours represent different sensors. 
    }
    \label{fig:data-seg-techniques}
\end{figure}

Activity recognition from passive sensors in smart homes is typically done in three steps.
First, the sparse incoming data is segmented using a windowing technique, second, a feature vector is constructed for the given window, and third, a classifier predicts the corresponding activity.
This section briefly reviews existing approaches for each of these steps.
Figure~\ref{fig:data-seg-techniques} shows an example sequence of activities with the associated sensor events and illustrates the two primary segmentation techniques: time-based and event-based windowing.

\subsection{Data Segmentation}
Time-based windowing divides the data stream into fixed-duration intervals, as used in~\cite{tapia2004activity, van2010activity}. 
This method is simple and works well with sensors that have a constant acquisition rate, like accelerometers and gyroscopes.
However, with passive sensors in smart homes, there may be large stretches of time without any events, hence methods need to be able to deal with empty data segments.
Using longer intervals can help, but leads to ambiguities when multiple shorter activities fall into the same window.

In event-based windowing, the data stream is divided into intervals with an equal number of sensor events, as in~\cite{krishnan2014activity, yala2015feature}. 
Typically, overlapping windows are used, where each sensor event triggers a new interval (see Figure~\ref{fig:data-seg-techniques}). 
The activity at the time of the latest sensor event within the window is used as a label for the whole window. This approach ensures that each segment is non-empty and provides context from events leading up to the activity.

Although event-based windowing is generally preferred for sparse streams from passive sensors, it also has drawbacks.
For instance, when a new activity starts after a long period of inactivity, previous events tend to be unrelated but still fall into the same window.
This is tackled in~\cite{krishnan2014activity} by dynamically estimating the optimal window size for each new sensor event, based on duration of activities and frequencies of sensor events.
Our work directly addresses this issue by proposing new features describing the temporal relationship between sensor events.

\subsection{Feature Extraction}
Given a segment with a certain number of events, a feature vector needs to be extracted to describe the segment.
A bag-of-words approach is used as a baseline in~\cite{krishnan2014activity}, capturing the frequencies of how many times each available sensor has fired during that window.
This approach naturally disregards any relationship between the sensors, such as spatial or temporal order.
% SWMI and its extensions
Sensor Window Mutual Information (SWMI) was proposed in~\cite{krishnan2014activity} to account for spatial context of sensors in the house.
The mutual information matrix indicates the likelihood of two sensors firing consecutively and is computed at training time.
During inference, it is then multiplied with the event frequencies vector to capture the contributions based on mutual information.
This method has been further extended to capture broader spatial dependencies during construction of the mutual information matrix~\cite{yala2015feature}.
Furthermore, \cite{ferretti2016experimental} extended SWMI by taking into account not just the sensor event co-occurrence, but also linking it to the context of a specific activity, effectively reducing the spurious influence of unrelated sensor events.
The sensors generally also communicate a status, which is usually binary.
In~\cite{yala2015feature}, the final state of the sensor is used instead of the event frequency as feature.
If a certain sensor would not have had events within the period, it is marked as $0$.

Finally, to incorporate the temporal order in which sensor events are triggered, Sensor Windows Time Weighting (SWTW)~\cite{krishnan2014activity} has been proposed, which puts a higher weight onto more recent events within the window.
A tabular representation of the methods referenced can be seen in Table~\ref{tab:competitors}.

We believe that incorporating further temporal cues is promising, as users engage in different activities at different times of day, and the exact timing of sensor events can provide crucial information about the activity.

\subsection{Classification}
After the feature vector is extracted from the data segment, it can be associated with the corresponding activity.
Knowledge-driven approaches use domain knowledge to model each activity and what pattern of sensor events they would trigger.
These activity templates can then be used for comparison with the observed feature vector~\cite{okeyo2014dynamic, rawashdeh2020knowledge, sfar2019dataseg}. 
However, this approach is intractable for a smart home setting, as it would require manual creation of activity templates for each user in their respective home.
Instead, most existing approaches are data-driven, and commonly used classifiers include Support Vector Machines (SVM)~\cite{krishnan2014activity, yala2015feature, ferretti2016experimental, sfar2019dataseg}, Naive Bayes~\cite{tapia2004activity}, and K-Nearest-Neighbour (KNN)~\cite{yala2017towards} classifier.

\section{Method} \label{sec:methodology}
This section presents our three main contributions.
Firstly, we extend SWMI by multiplying the feature vector with different weighting matrices depending on the time of day.
Secondly, we propose additional features to encode temporal cues, and thirdly, features encoding changes in location.

\subsection{Temporal Mutual Information} \label{sec:temp-swmi}
Mutual information (MI) is a valuable technique for capturing the spatial relationships between sensors in a smart home. However, traditional MI calculations treat these relationships as static, failing to account for the dynamic nature of human behaviour and its dependence on time. We address this limitation by introducing a temporal context-aware MI approach.

\subsubsection{Temporal Segmentation of the Day:}
We first segment the day into distinct periods reflecting the patterns of human activity throughout the day. Recognizing that activities often cluster around specific times, we divide the day into three categories: morning, afternoon, and night.

To determine the optimal boundaries, we employ a data-driven approach. We select potential thresholds to segment the day and cluster the activities in three groups. Then we calculate the cohesion of the activity patterns within each group as the the average distance from each point to the centroid of the cluster. We choose the one that minimizes this value, as lower values indicate tighter clusters. 
For each activity performed, we can define an activity-descriptor vector $\mathbf{a}_i$, as a one-hot encoding of the activity it represents, the hour $(\mathbf{a}_i)_h$ at which that activity starts, and the number of events that describe that activity.
Let $\mu$, $\alpha$, $\nu$ be the thresholds that define the morning, afternoon and night, such that $ 0 \leq \mu < \alpha < \nu < 24$, we can define the three groups of activities as follows: 
\begin{align}
    T_{\mu} &= \{\textbf{a}_i : (\mathbf{a}_i)_h \in [\mu, \alpha) \} \\
    T_{\alpha} &= \{\textbf{a}_i :  (\mathbf{a}_i)_h \in [\alpha, \nu) \} \\
    T_{\nu} &= \{\textbf{a}_i : (\mathbf{a}_i)_h \in [\nu, 24) \cup \ [0, \mu) \}
\end{align}
Now, we formulate a minimization problem to obtain the $\mu$, $\alpha$, $\nu$ that give the most cohesive activity patterns.
This can be interpreted as building temporal clusters with similar activities.
Equation~\ref{eq:cohesion} computes the average distance from the cluster centroid~$\bar{\textbf{a}}_{T_\chi}$ for each group $T_\chi \in \{T_{\mu}, T_{\alpha}, T_{\nu}\}$, and then minimizes the mean of the three groups:
\begin{equation} \label{eq:cohesion}
  \argmin_{\mu, \alpha, \nu} \, \frac{1}{3} \, \sum_{T_\chi \in \{T_\mu, T_\alpha, T_\nu\}} \hspace{3pt}\left( \frac{1}{|T_\chi|} \sum_{\textbf{a}_i \in T_\chi} \|\textbf{a}_i- \bar{\textbf{a}}_{T_\chi}\|^2 \right)
\end{equation}

\subsubsection{Refined MI Calculation:}
We refine the MI calculation by incorporating this temporal segmentation. Instead of computing a global MI matrix for all sensor pairs, we calculate separate MI matrices for morning, afternoon, and night. This allows us to capture the varying dependencies between sensors during different parts of the day. 

A sensor event $\textbf{e}_i$ is defined by the timestamp at which the event occurs, the binary sensor status (ON/OFF) and the sensor location $s_i\in \{S_1, ..., S_M\}$ with $M$ the number of sensors deployed in the house. So for $T_\chi \in \{T_\mu, T_\alpha, T_\nu\}$, we can define the mutual information matrix for that segment of the day as:

\begin{equation} \label{eq:swmi_1}
    MI_{T_\chi}(i, j) = \frac{1}{|T_\chi|} \sum_{t=1}^{|T_\chi|-1} \delta(s_t, S_i) \cdot \delta(s_{t+1}, S_j) 
\end{equation}
where
\begin{equation}\label{eq:swmi_2}
\delta\left(s_t, S_i\right)= \begin{cases}
        0 & \text { if } s_t \neq S_i \\ 
        1 & \text { if } s_t=S_i
        \end{cases}
\end{equation}

The matrices are defined offline, and used as weights while constructing the feature vector, as shown in Figure \ref{fig:process}. Each event of the sequence is weighted with respect to the last event in the window, utilizing its respective matrix according to the time of day.

\subsection{Expanded Feature Space} \label{sec:feature-space}
In addition, we expand the feature space to explicitly incorporate temporal context. Cyclical time encoding addresses the limitation of acyclic time encoding, where consecutive values (e.g., 23:00 and 00:00) might incorrectly appear far apart in the feature space due to their numerical distance.

\subsubsection{Cyclical Time Encoding:}
\begin{figure}[h]
    \centering
    \begin{tabular}{ccc}
        \includegraphics[height=4.45cm]{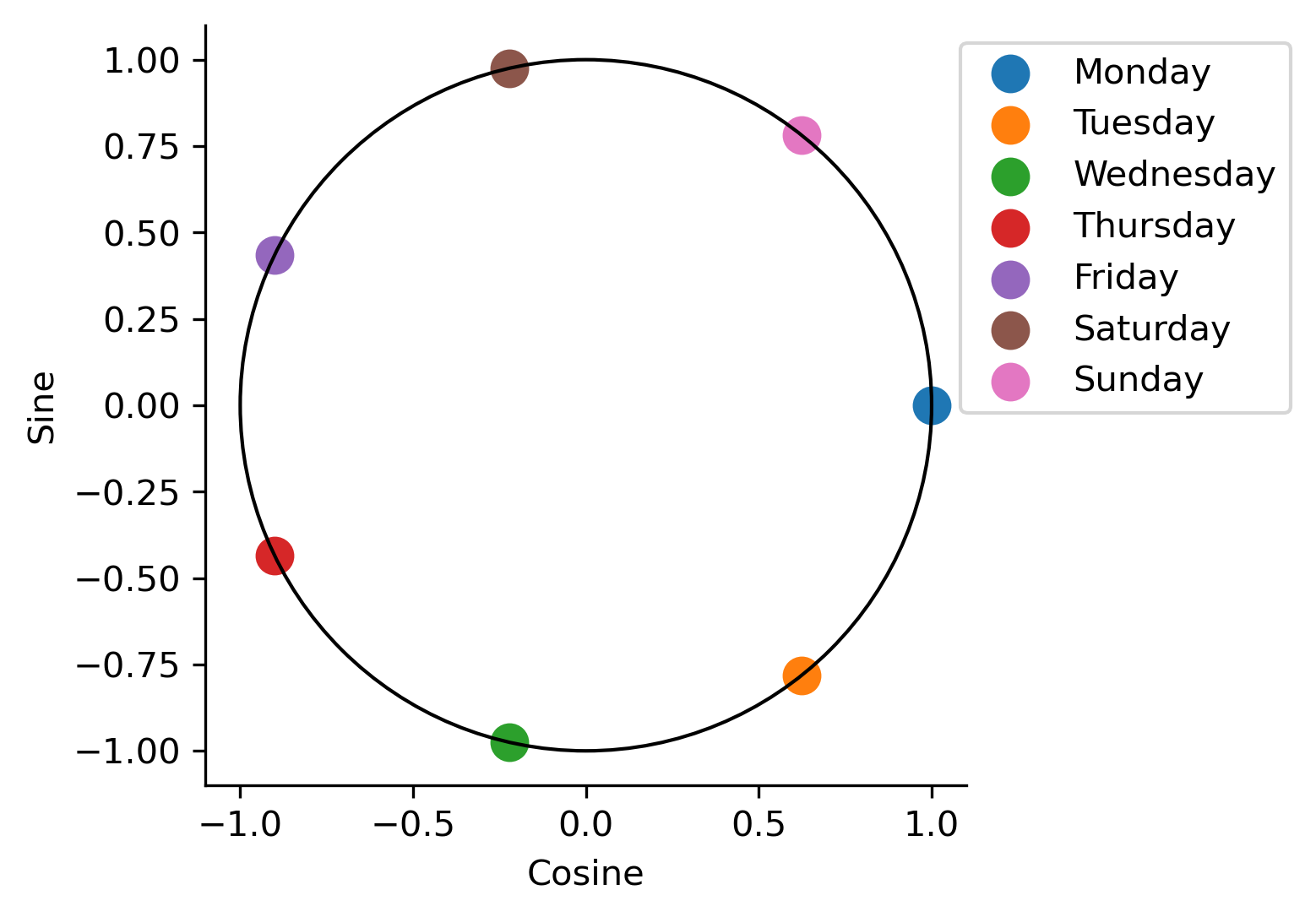} & \hspace{0.2cm} &
         \includegraphics[height=4.45cm]{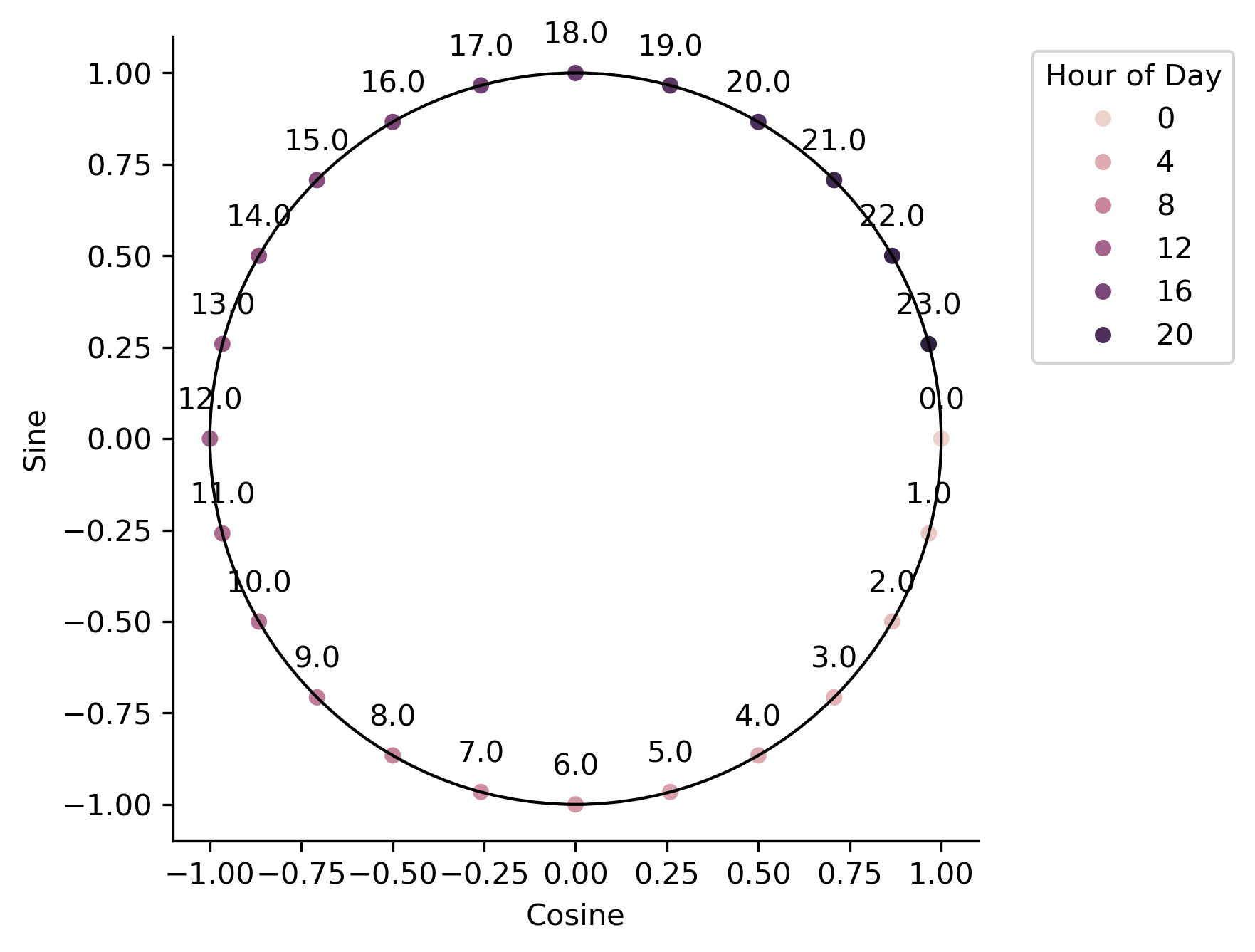}
    \end{tabular}
    \caption{Visualization of the day of the week and hour of the day through sine and cosine transformations to capture their cyclical nature. This ensures that ``Monday'' is close to both ``Sunday'' and ``Tuesday'' in the feature space.}
    \label{fig:sin-cos}
\end{figure}

We represent the hour of the day ($h$) and the day of the week ($d$) using sine and cosine functions to capture their cyclical nature:
\begin{align}
\text{Hour of Day (sin): } & \sin (2\pi * h/24) \\
\text{Hour of Day (cos): } & \cos (2\pi * h/24) \\
\text{Day of Week (sin): } & \sin (2\pi * d/7) \\
\text{Day of Week (cos): } & \cos (2\pi * d/7)
\end{align}
We can visualize these features in Figure \ref{fig:sin-cos}. This encoding allows the KNN classifier to better understand the relationships between different hours and days, as similar times will have closer representations in the feature space.

\subsubsection{Location Change Feature:}
We introduce a binary feature to indicate when a change in location occurs, i.e. if the last event of the sequence is coming from a different sensor than the antecedent. This feature could signal a potential shift in activity and it would help to capture the transition points between activities, providing additional context to the KNN classifier.

\section{Experiments and Results} \label{sec:experiments}

In the following subsections we explain the evaluation procedure and then present and discuss the results.

\subsection{Evaluation Procedure}

We evaluated our approach using four real-world smart home datasets, part of the Center for Advanced Studies in Adaptive Systems (CASAS) project \cite{cook2009collecting, cook2012casas}:
\begin{itemize}
    \item \textbf{TM Datasets (TM006, TM007, TM008)}: 
    These datasets contain sensor data from single-resident apartments where participants installed the ``Smart Home in a Box'' kit themselves.
     The datasets include data from 17-19~binary sensors in 7-8~locations (rooms) within the apartments, spanning 30~days of labelled activities.
     In the publicly available version of the datasets, 
     the sensors are mapped to the room they are in.
     Hence, we cannot differentiate between sensors that are in the same room.
    We chose these datasets due to their real-world nature and the practical implications for telecare deployments. 
     
    \item \textbf{Aruba Dataset}: This dataset captures sensor data from a smart home occupied by an elderly woman, regularly visited by her children and grandchildren. We consider the first three months of this dataset, containing data from 39 sensors, including motion detectors, door sensors, and temperature sensors. 
    Although temperature sensors are not binary sensors, in this case, we use them as such, i.e., just their activation.
    The Aruba dataset is a common benchmark for evaluating data segmentation techniques \cite{wan2015dynamic, yala2015feature, machot2017activity, sfar2019dataseg, najeh2022dynamic}.
\end{itemize}

All datasets are manually labelled, annotating basic and instrumental ADLs like bathing, cooking, eating, sleeping, entering/leaving the home, etc.  To reduce the overall number of activity classes and facilitate analysis, we aggregated similar classes (\textit{e.g.}, ``Cook breakfast'', ``Cook lunch'', and ``Cook dinner'' are combined into ``Cook''). 
 These datasets are imbalanced, with the ``Other'' activity class representing 46.4\% to 49.5\% of recorded events.

We evaluate the performance of our approach using two key metrics: accuracy, measuring the overall proportion of correctly classified sensor events, and F1-score, which, as the harmonic mean of precision and recall, is particularly useful for imbalanced datasets. In this multi-class problem, we report the weighted-averaged F1-score, i.e. the mean of all per-class F1-scores, weighted by the support of each class. We report both metrics for two cases: first, including all classes and second, excluding the ``Other'' class.
The latter is useful as the ``Other'' class is predominant, but less meaningful.

We adopted a K-Nearest Neighbours (KNN) classifier for these experiments. KNN is known for its simplicity, flexibility, and ability to handle multi-class datasets without needing a dedicated training phase.
The data is split temporally in training, validation and test set. 
Hyperparameter tuning has been performed separetely for each user on the respective validation set.
Both $K$ and the window size were fine-tuned through grid-search for each method that has been evaluated. 
The results presented in the following section are on the test set, after retraining the model on the whole training and validation set.

\subsection{Results and Discussion}
To evaluate our contributions proposed in Section~\ref{sec:methodology}, we compare them with  a range of existing state-of-the-art methods (see Table~\ref{tab:competitors}).

\begin{table}[htb]
    \centering
    \caption{Online activity recognition methods based on passive sensors and using sensor-based data segmentation.}
    \begin{tabular}
    {>{\centering\arraybackslash}p{1.6cm}| >{\arraybackslash}p{10.3cm} }
        Method & Description \\ \hline \hline
        SW & Uses sensor event-based windowing with feature vectors as counts of sensor activations per window \cite{krishnan2014activity}. \\
    SWMI & Weighs sensor readings by mutual information relative to the last event, based on the probability of consecutive sensor firings \cite{krishnan2014activity}. \\
    SWMIex & Extends SWMI by defining mutual information as the likelihood of sensors occurring together within a window \cite{yala2015feature, yala2017towards}. \\
    SWMI Act & Computes mutual information within sensor sequences for predefined activities, excluding ``Other'' activity \cite{ferretti2016experimental}. \\
    BSS & The last state of each sensor in a window is used as weight: 1 for ``ON'', -1 for ``OFF'', and 0 for absence  \cite{ferretti2016experimental}. \\
    SWLS & Incorporates the final state of binary sensors, with ``ON'' as +1 and ``OFF'' as -1 \cite{yala2015feature}. \\
    DW & Uses a probabilistic approach to determine optimal window size based on activity duration and sensor event frequencies \cite{krishnan2014activity}. \\
    SWTW & Contribution of event to feature vector is weighted since the time passed from last event \cite{krishnan2014activity}. \\
\hline
    \end{tabular}
    \label{tab:competitors}
\end{table}

We proposed three distinct contributions.
While the temporal mutual information can be considered an extension of SWMI~\cite{krishnan2014activity}, our cyclical time-encoding features and location-change features are compatible with any method.
We therefore test the following variants to assess the performance gain by the individual contributions:

\begin{itemize}
    \item Cyclical time-encoding features, described in Section~\ref{sec:feature-space} are appended to a SWMI~\cite{krishnan2014activity} base vector and to a BSS~\cite{ferretti2016experimental} base vector. 
    \item Location-change feature, presented in Section~\ref{sec:feature-space}, is appended to SWMI base vector.
    \item Combining both the cyclical time-encoding features and the location-change feature to a SWMI base vector.
    \item SWMI-Temp, introduced in Section \ref{sec:temp-swmi}, weighs the feature vector using separate SWMI matrices for morning, afternoon, night.
    \item Combined: SWMI-Temp plus cyclical time-encoding and location-change features.
\end{itemize}

\begin{table}[h]
    \centering
    \caption{Performance metrics table for TM006, TM007, TM008 and Aruba. The best
results are highlighted in bold, while the second-best results are underlined}
      \begin{tabular}{ >{\centering\arraybackslash}m{2.6cm}| >{\centering\arraybackslash}m{1.09cm} >{\centering\arraybackslash}m{1.09cm}|>{\centering\arraybackslash}m{1.09cm}>{\centering\arraybackslash}m{1.09cm}|>{\centering\arraybackslash}m{1.09cm}>{\centering\arraybackslash}m{1.09cm}|>{\centering\arraybackslash}m{1.09cm}>{\centering\arraybackslash}m{1.09cm}}
% \hline
                  & \multicolumn{2}{c|}{TM006}     &  \multicolumn{2}{c|}{TM007}    & \multicolumn{2}{c|}{TM008}  & \multicolumn{2}{c}{Aruba} \\
                  & Acc. & F1 & Acc. & F1 & Acc. & F1 & Acc. & F1 \\ \hline \hline
    SW \cite{krishnan2014activity}      & 0.537 & 0.446 & 0.570 & 0.533  & 0.502 & 0.464 & 0.774 & 0.763 \\
    BSS \cite{ferretti2016experimental}          & 0.553 & 0.470 & 0.563 & 0.533  & 0.460 & 0.412 & \underline{0.775} & \underline{0.765} \\
    DW \cite{krishnan2014activity}           & 0.483 & 0.456 & 0.589 & 0.573  & 0.450 & 0.392 & 0.755 & 0.745 \\
    SWLS \cite{yala2015feature}         & 0.554 & 0.466 & 0.557 & 0.531  & 0.489 & 0.455 & 0.765 & 0.756 \\
     SWMI \cite{krishnan2014activity}         & 0.577 & 0.546 & 0.604 & 0.594  & 0.618 & 0.614 & 0.761 & 0.750 \\
    SWMI Act \cite{ferretti2016experimental}     & 0.576 & 0.547 & 0.610 & 0.601  & 0.596 & 0.593 & 0.761 & 0.749 \\
    SWMIex \cite{yala2015feature}     & 0.576 & 0.545 & 0.588 & 0.560  & 0.608 & 0.603 & 0.768 & 0.757 \\
    SWTW \cite{krishnan2014activity}         & 0.491 & 0.464 & 0.605 & 0.606  & 0.556 & 0.569 & 0.757 & 0.746 \\ \hline 
    BSS + cyclic features & 0.601 & 0.578 & 0.597 & 0.584 & 0.662 & 0.663 & \textbf{0.777} & \textbf{0.766} \\
    SWMI + cyclic features & \underline{0.638} & \underline{0.626} & 0.646 & \underline{0.634} & 0.706 & 0.703 & 0.757 & 0.744 \\
    SWMI + location change feature & 0.594  & 0.568 & 0.631 & 0.612 & 0.619 & 0.615 & 0.764 & 0.752 \\
    SWMI + cyclic + location  & \textbf{0.644} & \textbf{0.631} & \textbf{0.652} & \textbf{0.640} & \textbf{0.712} & \textbf{0.710} & 0.764 & 0.753 \\
    SWMI-Temp & 0.615 & 0.565 & 0.616 & 0.594 & 0.642 & 0.633 & 0.762 & 0.752 \\
    Combined     & 0.632 & 0.620 & \underline{0.652} & 0.625 & \underline{0.707} & \underline{0.705} & 0.763 & 0.750 \\
    \hline
    \end{tabular}
    \label{tab:results}
\end{table}

\begin{table}[h]
    \centering
    \caption{Performance metrics table for TM006, TM007, TM008 and Aruba, excluding ``Other'' class from evaluation. The best
results are highlighted in bold, while the second-best results are underlined}
   \begin{tabular}{ >{\centering\arraybackslash}m{2.6cm}| >{\centering\arraybackslash}m{1.09cm} >{\centering\arraybackslash}m{1.09cm}|>{\centering\arraybackslash}m{1.09cm}>{\centering\arraybackslash}m{1.09cm}|>{\centering\arraybackslash}m{1.09cm}>{\centering\arraybackslash}m{1.09cm}|>{\centering\arraybackslash}m{1.09cm}>{\centering\arraybackslash}m{1.09cm}}
% \hline
                   & \multicolumn{2}{c|}{TM006}     &  \multicolumn{2}{c|}{TM007}    & \multicolumn{2}{c|}{TM008}  & \multicolumn{2}{c}{Aruba} \\
                  & Acc. & F1 & Acc. & F1 & Acc. & F1 & Acc. & F1 \\ \hline \hline

    SW \cite{krishnan2014activity}      & 0.251 & 0.233 & 0.331 & 0.316 & 0.261 & 0.333 & 0.696 & 0.785 \\
    BSS \cite{ferretti2016experimental}          & 0.274 & 0.277 & 0.345 & 0.326 & 0.251 & 0.282 & \textbf{0.718} & \textbf{0.796} \\
    DW  \cite{krishnan2014activity}          & 0.337 & 0.337 & 0.347 & 0.381 & 0.240 & 0.251 & 0.653 & 0.749 \\
    SWLS  \cite{yala2015feature}        & 0.251 & 0.273 & 0.347 & 0.343 & 0.274 & 0.327 & \underline{0.711} & 0.789 \\
    SWMI \cite{krishnan2014activity}         & 0.337 & 0.398 & 0.516 & 0.545 & 0.525 & 0.579 & 0.667 & 0.764 \\
    SWMI Act \cite{ferretti2016experimental}     & 0.344 & 0.405 & \underline{0.533} & \underline{0.560} & 0.596 & 0.593 & 0.653 & 0.755 \\
    SWMIex \cite{yala2015feature}     & 0.336 & 0.401 & 0.378 & 0.372 & 0.527 & 0.584 & 0.680 & 0.772 \\
    SWTW  \cite{krishnan2014activity}        & 0.293 & 0.316 & 0.504 & 0.522 & 0.515 & 0.569 & 0.687 & 0.773 \\ \hline
    BSS + cyclic features & 0.403 & 0.468 & 0.519 & 0.545 & 0.631 & 0.677 & 0.705 & \underline{0.790} \\
    SWMI + cyclic features & \underline{0.447}  & \underline{0.512}  & 0.513  & 0.543  &  \underline{0.640} & \underline{0.708} & 0.643 & 0.748 \\
    SWMI + location change feature  & 0.348 & 	0.412 & \textbf{0.559} & \textbf{0.571} & 0.504 & 0.559  & 0.664 &	0.764 \\ 
    SWMI + cyclic + location  & \textbf{0.451} & \textbf{0.523} & 0.516 & 0.545 & \textbf{0.649} & \textbf{0.715} & 0.653  &  0.757\\
    SWMI-Temp  & 0.383 & 0.432 & 0.429 & 0.462 & 0.516 & 0.576 & 0.663 & 0.763 \\
    Combined     & 0.440 & 0.508 & 0.492 & 0.530 & 0.633 & 0.686 & 0.628 & 0.742 \\
    \hline
    \end{tabular}
    \label{tab:results-no-other}
\end{table}

Table~\ref{tab:results} presents the results of the experiments performed including all the classes, and Table~\ref{tab:results-no-other} excludes the ``Other'' class. 
In both tables, the upper part presents the results of the state-of-the-art methods, while the lower section of the table presents the results of our contributions. 

In almost all cases, our contributions consistently improve the scores of their respective baseline method, but the margin of improvement varies.
Looking at Table~\ref{tab:results}, the best results for TM users are achieved by combining SWMI with both the cyclical time-encoding features and the location-change feature, surpassing the state-of-the-art methods by a large margin for both accuracy and F1-score.
Across all datasets, ``SWMI + cyclic + location'' gives an average improvement of 8.83\% in terms of accuracy and 9.83\% in terms of F1-score over the base SWMI method.

For the Aruba dataset, the BSS method with our cyclical time-encoding features performs best, but the improvement over the plain BSS method is marginal.
Adding the cyclic features to the BSS feature vector, we obtain an average improvement across all datasets of 5.91\% in terms of accuracy and 13.91\% in terms of F1-score.
When excluding the ``Other'' class (see Table~\ref{tab:results-no-other}), we obtain an average improvement of 35.70\% in terms of accuracy and 51.08\% in terms of F1-score, demonstrating that the performance gain indeed stems from meaningful detection of the relevant activities.

In both tables, we observe a significant difference in terms of performance between the TM datasets and Aruba.
The latter has a different structure compared to the TM users, both in terms of number of sensors and amount of data available.
Based on the results, it seems that the TM datasets benefit most from adding features for temporal context, while there are no consistent benefits on the Aruba dataset.
This indicates that the temporal context is particularly important for HAR in a low-data regime, as the TM datasets use fewer sensors.
In practice, this is crucial, as rolling out sensor kits for smart homes is cost-sensitive and being able to recognize activities from fewer sensors will significantly boost the economical viability of a large-scale implementation. 

Interestingly, there is no clear winning method across all datasets.
But even though the results are inconclusive in establishing a method that outperforms the others for all experiments, they clearly demonstrate the usefulness of adding temporal context to enhance the performance of human activity recognition.

\begin{figure}[h!]
    \centering
    % left bottom right top,
    \includegraphics[trim=0 0 0 5,clip,width=10cm]{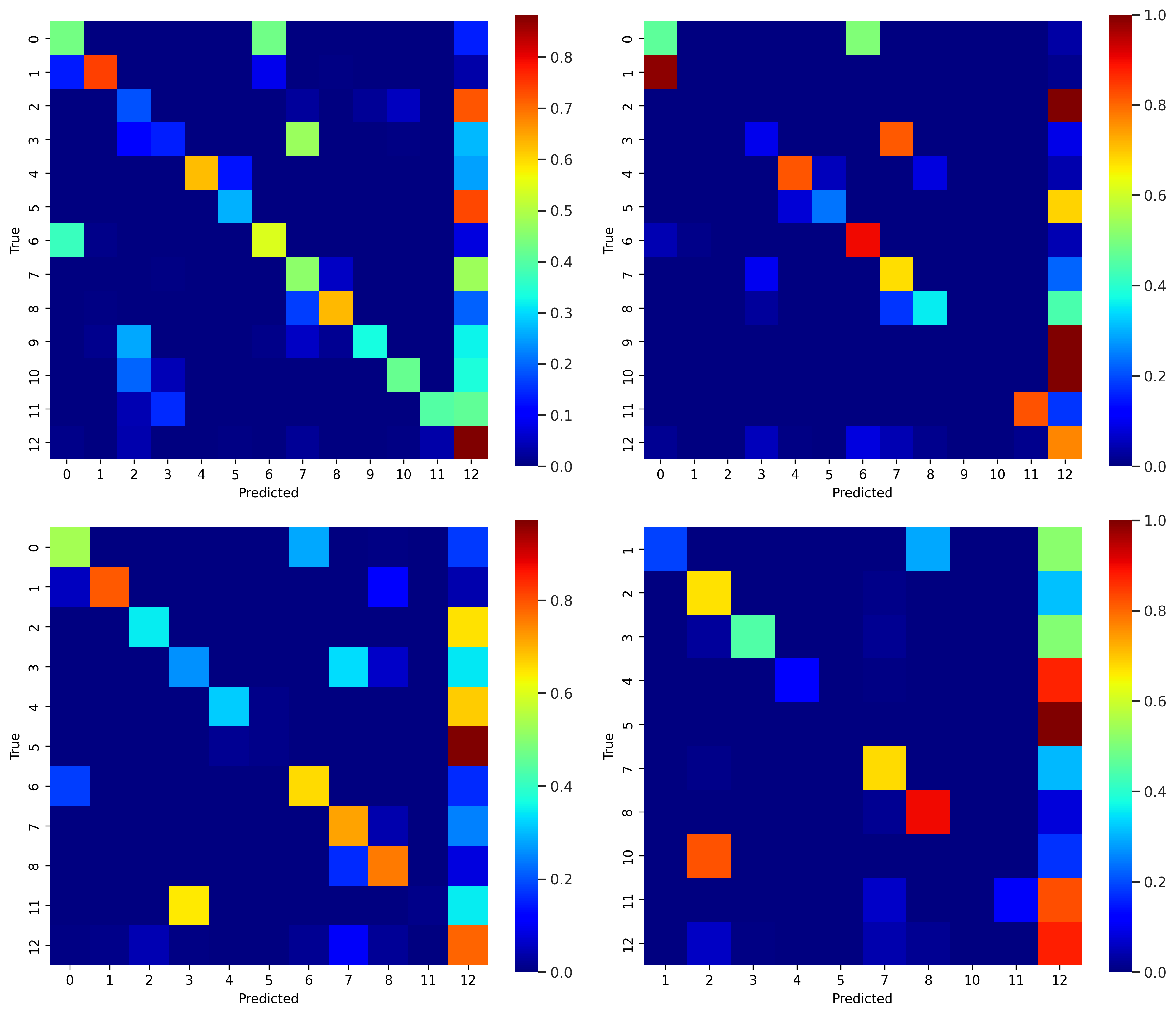}
    \caption{Normalized confusion matrix of test set results for the SWMI method with cyclic features and location change feature, for TM006 (top-left), TM007 (top-right), TM008 (bottom-left), and Aruba (bottom-right). The labels are: 0) Bathing, 1) Bed to toilet, 2) Cook / Meal Preparation, 3) Eat, 4) Enter home, 5) Leave home, 6) Personal Hygiene, 7) Relax, 8) Sleep, 9) Take Medicine, 10) Wash Dishes, 11) Work and 12) Other.}
    \label{fig:confusion-matrix}
\end{figure}
To gain some qualitative insights, we visualize the classification performance for different activities when using the SWMI method with cyclic features and location-change feature.
Figure~\ref{fig:confusion-matrix} shows the normalized confusion matrices for each of the datasets.
Activities like ``Personal Hygiene'', ``Relax'' and ``Sleep'' are consistently captured well, likely due to their spatial and temporal patterns.
In contrast, ``Leave Home'' and ``Enter Home'' are frequently missed, potentially due to variations in timing and sensor readings.
It is clearly visible that all classes are regularly confused with the ``Other'' class.
This class absorbs most of the events but is quite heterogeneous, which makes it difficult to distinguish it from other predefined activities.

\section{Conclusion and Future Works} \label{sec:conclusion}

This paper presented three distinct methods for integrating temporal context into Human Activity Recognition in smart homes, aiming to enhance the accuracy of activity detection for elderly individuals living independently. 
By recognizing the cyclical nature of human behaviour, the first approach segments the day into intervals and calculates distinct Mutual Information matrices for each period. 
Furthermore, we presented two new types of features, the cyclical time-encoding features and the location-change feature. Evaluating these contributions on real-world datasets demonstrated increased accuracy and F1-score compared with other state-of-the-art techniques on three out of four datasets.
In particular, when the spatial context provided by the sensors is reduced, HAR can truly benefit from the temporal context provided by the additional features. This insight is particular useful in telecare industry where the numbers of devices deployed is usually limited.
However, the research is limited to single-inhabitant apartments, and future work should explore multi-person environments to further validate and extend the methodology.
Moreover, the models developed in this work are highly user-specific. Future research should investigate how to generalize to new users, perhaps in a self-supervised or unsupervised manner utilizing unlabelled datasets.

The findings underscore the importance of temporal information in HAR systems and suggest potential advancements in smart home technologies for better supporting independent living for the elderly.


\bibliographystyle{splncs04}
\bibliography{bibliography}

@article{un-data-2022,
    author = {{United Nations, Department of Economic and Social Affairs, Population Division}},
    title = {World Population Prospects 2022, Data Sources.},
    journal = {UN DESA/POP/2022/DC/NO. 9.},
    year = {2022}
}

@book{world2015world,
  title={World report on ageing and health},
  author={{World Health Organization}},
  year={2015},
  publisher={World Health Organization}
}

@article{marek2012aging,
  title={Aging in place versus nursing home care: Comparison of costs to Medicare and Medicaid},
  author={Marek, Karen Dorman and Stetzer, Frank and Adams, Scott J and Popejoy, Lori L and Rantz, Marilyn},
  journal={Research in gerontological nursing},
  volume={5},
  number={2},
  pages={123--129},
  year={2012},
  publisher={Slack Incorporated Thorofare, NJ}
}

@article{perry2014relocation,
  title={Relocation remembered: Perspectives on senior transitions in the living environment},
  author={Perry, Tam E and Andersen, Troy C and Kaplan, Daniel B},
  journal={The Gerontologist},
  volume={54},
  number={1},
  pages={75--81},
  year={2014},
  publisher={Oxford University Press US}
}

@article{golant2008commentary,
  title={Commentary: Irrational exuberance for the aging in place of vulnerable low-income older homeowners},
  author={Golant, Stephen M},
  journal={Journal of Aging \& Social Policy},
  volume={20},
  number={4},
  pages={379--397},
  year={2008},
  publisher={Taylor \& Francis}
}

@article{xu2022online,
  title={Online Activity Recognition Combining Dynamic Segmentation and Emergent Modeling},
  author={Xu, Zimin and Wang, Guoli and Guo, Xuemei},
  journal={Sensors},
  volume={22},
  number={6},
  pages={2250},
  year={2022},
  publisher={MDPI}
}

@article{van2010activity,
  title={An activity monitoring system for elderly care using generative and discriminative models},
  author={Van Kasteren, TLM and Englebienne, Gwenn and Kr{\"o}se, Ben JA},
  journal={Personal and ubiquitous computing},
  volume={14},
  pages={489--498},
  year={2010},
  publisher={Springer}
}

@article{wang2023survey,
  title={A Survey on Ambient Sensor-Based Abnormal Behaviour Detection for Elderly People in Healthcare},
  author={Wang, Yan and Wang, Xin and Arifoglu, Damla and Lu, Chenggang and Bouchachia, Abdelhamid and Geng, Yingrui and Zheng, Ge},
  journal={Electronics},
  volume={12},
  number={7},
  pages={1539},
  year={2023},
  publisher={MDPI}
}

@article{deep2019survey,
  title={A survey on anomalous behavior detection for elderly care using dense-sensing networks},
  author={Deep, Samundra and Zheng, Xi and Karmakar, Chandan and Yu, Dongjin and Hamey, Leonard GC and Jin, Jiong},
  journal={IEEE Communications Surveys \& Tutorials},
  volume={22},
  number={1},
  pages={352--370},
  year={2019},
  publisher={IEEE}
}

@article{edemekong2019activities,
  title={Activities of daily living},
  author={Edemekong, Peter F and Bomgaars, Deb and Sukumaran, Sukesh and Levy, Shoshana B},
  year={2019},
  publisher={StatPearls Publishing LLC}
}

@INPROCEEDINGS{das2012prompt,
  author={Das, Barnan and Seelye, Adriana M. and Thomas, Brian L. and Cook, Diane J. and Holder, Larry B. and Schmitter-Edgecombe, Maureen},
  booktitle={2012 IEEE Consumer Communications and Networking Conference (CCNC)}, 
  title={Using smart phones for context-aware prompting in smart environments}, 
  year={2012},
  volume={},
  number={},
  pages={399-403},
  doi={10.1109/CCNC.2012.6181023}
}

@inproceedings{das2011automated,
  title={An automated prompting system for smart environments},
  author={Das, Barnan and Chen, Chao and Seelye, Adriana M and Cook, Diane J},
  booktitle={Toward Useful Services for Elderly and People with Disabilities: 9th International Conference on Smart Homes and Health Telematics, ICOST 2011, Montreal, Canada, June 20-22, 2011. Proceedings 9},
  pages={9--16},
  year={2011},
  organization={Springer}
}

@article{cook2012casas,
  title={CASAS: A smart home in a box},
  author={Cook, Diane J and Crandall, Aaron S and Thomas, Brian L and Krishnan, Narayanan C},
  journal={Computer},
  volume={46},
  number={7},
  pages={62--69},
  year={2012},
  publisher={IEEE}
}

@article{zhang2022deep,
  title={Deep learning in human activity recognition with wearable sensors: A review on advances},
  author={Zhang, Shibo and Li, Yaxuan and Zhang, Shen and Shahabi, Farzad and Xia, Stephen and Deng, Yu and Alshurafa, Nabil},
  journal={Sensors},
  volume={22},
  number={4},
  pages={1476},
  year={2022},
  publisher={MDPI}
}

@article{dang2020sensor,
  title={Sensor-based and vision-based human activity recognition: A comprehensive survey},
  author={Dang, L Minh and Min, Kyungbok and Wang, Hanxiang and Piran, Md Jalil and Lee, Cheol Hee and Moon, Hyeonjoon},
  journal={Pattern Recognition},
  volume={108},
  pages={107561},
  year={2020},
  publisher={Elsevier}
}

@article{claes2015attitudes,
  title={Attitudes and perceptions of adults of 60 years and older towards in-home monitoring of the activities of daily living with contactless sensors: an explorative study},
  author={Claes, Veerle and Devriendt, Els and Tournoy, Jos and Milisen, Koen},
  journal={International journal of nursing studies},
  volume={52},
  number={1},
  pages={134--148},
  year={2015},
  publisher={Elsevier}
}

@article{krishnan2014activity,
  title={Activity recognition on streaming sensor data},
  author={Krishnan, Narayanan C and Cook, Diane J},
  journal={Pervasive and mobile computing},
  volume={10},
  pages={138--154},
  year={2014},
  publisher={Elsevier}
}

@inproceedings{yala2015feature,
  title={Feature extraction for human activity recognition on streaming data},
  author={Yala, Nawel and Fergani, Belkacem and Fleury, Anthony},
  booktitle={2015 International symposium on innovations in intelligent systems and applications (INISTA)},
  pages={1--6},
  year={2015},
  organization={IEEE}
}

@inproceedings{ferretti2016experimental,
  title={An experimental study on new features for activity of daily living recognition},
  author={Ferretti, Daniele and Principi, Emanuele and Squartini, Stefano and Mandolini, Luigi},
  booktitle={2016 International Joint Conference on Neural Networks (IJCNN)},
  pages={3958--3965},
  year={2016},
  organization={IEEE}
}

@article{okeyo2014dynamic,
  title={Dynamic sensor data segmentation for real-time knowledge-driven activity recognition},
  author={Okeyo, George and Chen, Liming and Wang, Hui and Sterritt, Roy},
  journal={Pervasive and Mobile Computing},
  volume={10},
  pages={155--172},
  year={2014},
  publisher={Elsevier}
}

@article{rawashdeh2020knowledge,
  title={A knowledge-driven approach for activity recognition in smart homes based on activity profiling},
  author={Rawashdeh, Majdi and Al Zamil, Mohammed GH and Samarah, Samer and Hossain, M Shamim and Muhammad, Ghulam},
  journal={Future Generation Computer Systems},
  volume={107},
  pages={924--941},
  year={2020},
  publisher={Elsevier}
}

@inproceedings{sfar2019dataseg,
  title={DataSeg: Dynamic streaming sensor data segmentation for activity recognition},
  author={Sfar, Hela and Bouzeghoub, Amel},
  booktitle={Proceedings of the 34th ACM/SIGAPP Symposium on Applied Computing},
  pages={557--563},
  year={2019}
}

@article{yala2017towards,
  title={Towards improving feature extraction and classification for activity recognition on streaming data},
  author={Yala, Nawel and Fergani, Belkacem and Fleury, Anthony},
  journal={Journal of Ambient Intelligence and Humanized Computing},
  volume={8},
  number={2},
  pages={177--189},
  year={2017},
  publisher={Springer}
}

@inproceedings{cook2009collecting,
  title={Collecting and disseminating smart home sensor data in the CASAS project},
  author={Cook, Diane and Schmitter-Edgecombe, Maureen and Crandall, Aaron and Sanders, Chad and Thomas, Brian},
  booktitle={Proceedings of the CHI workshop on developing shared home behavior datasets to advance HCI and ubiquitous computing research},
  pages={1--7},
  year={2009},
  organization={IEEE}
}

@inproceedings{tapia2004activity,
  title={Activity recognition in the home using simple and ubiquitous sensors},
  author={Tapia, Emmanuel Munguia and Intille, Stephen S and Larson, Kent},
  booktitle={International conference on pervasive computing},
  pages={158--175},
  year={2004},
  organization={Springer}
}

@article{wan2015dynamic,
  title={Dynamic sensor event segmentation for real-time activity recognition in a smart home context},
  author={Wan, Jie and O’grady, Michael J and O’Hare, Gregory MP},
  journal={Personal and Ubiquitous Computing},
  volume={19},
  pages={287--301},
  year={2015},
  publisher={Springer}
}

@inproceedings{intille2005living,
  title={A living laboratory for the design and evaluation of ubiquitous computing technologies},
  author={Intille, Stephen S and Larson, Kent and Beaudin, Jennifer S and Nawyn, Jason and Tapia, E Munguia and Kaushik, Pallavi},
  booktitle={CHI'05 extended abstracts on Human factors in computing systems},
  pages={1941--1944},
  year={2005}
}

@article{machot2017activity,
  title={Activity recognition in sensor data streams for active and assisted living environments},
  author={Al Machot, Fadi and Mosa, Ahmad Haj and Ali, Mouhannad and Kyamakya, Kyandoghere},
  journal={IEEE Transactions on Circuits and Systems for Video Technology},
  volume={28},
  number={10},
  pages={2933--2945},
  year={2017},
  publisher={IEEE}
}

@article{najeh2022dynamic,
  title={Dynamic segmentation of sensor events for real-time human activity recognition in a smart home context},
  author={Najeh, Houda and Lohr, Christophe and Leduc, Benoit},
  journal={Sensors},
  volume={22},
  number={14},
  pages={5458},
  year={2022},
  publisher={MDPI}
}
\end{document}